\documentclass[conference,a4paper]{IEEEtran}
\usepackage{iftex}
\ifPDFTeX
  \usepackage[T1]{fontenc}
\else
  \usepackage{fontspec}
  \usepackage{unicode-math}
\fi
\usepackage[a4paper,left=0.6in,right=0.6in,top=1in,bottom=1in,footskip=32pt]{geometry}
\usepackage{amsmath}
\ifPDFTeX\usepackage{amssymb}\fi
\usepackage{graphicx}
\usepackage{array}
\usepackage{booktabs}
\usepackage{xcolor}
\usepackage{url}
\newcolumntype{P}[1]{>{\raggedright\arraybackslash}p{#1}}
\definecolor{tgt}{RGB}{31,94,166}
\definecolor{rnd}{RGB}{150,150,150}
\definecolor{bad}{RGB}{214,110,30}
\usepackage{fancyhdr}
\fancypagestyle{correspondingauthor}{%
  \fancyhf{}%
  \fancyfoot[L]{\footnotesize\textsuperscript{*} Corresponding author}%
}
\renewcommand{\IEEEkeywordsname}{Keywords}
\begin{document}

\title{LM$^2$: Compiling a Language Model into Another Language Model\textquoteright s Weights}

\author{\IEEEauthorblockN{Naman Chawla}
\IEEEauthorblockA{\textit{Bhagwan Parshuram Institute of Technology}\\
\textit{Guru Gobind Singh Indraprastha University}\\
Delhi, India\\
namanchawla2404@gmail.com}
\and
\IEEEauthorblockN{Chhavi Dhiman\textsuperscript{*}}
\IEEEauthorblockA{\textit{Delhi Technological University}\\
Delhi, India\\
chhavi.dhiman@dtu.ac.in}}

\maketitle
\thispagestyle{correspondingauthor}

\begin{abstract}
Deploying a conversational model and a specialist together could avoid maintaining a separate specialist service. We present LM$^2$, which embeds a trained story generator into a larger conversational model. The larger model handles the conversation and invokes the generator for story requests. We constrain the generator's arithmetic during training so its calculations can be compiled directly into added model layers. We compare this approach with QwenPU, which executes the stories260K language model on an embedded processor. LM$^2$ requires one host token per generated character, compared with approximately 25{,}000 for QwenPU. However, its story generator predicts held-out text less well: 1.145 bits per character versus 0.894 for stories260K, where lower is better. The Qwen3.5-0.8B implementation occupies 18.1 GB and generates the same characters as a standalone integer implementation of the inner model in the tested runs. These results illustrate a trade-off between host-token cost and text-prediction performance.
\end{abstract}


\section*{Nomenclature}
\noindent{\footnotesize
\begin{tabular}{@{}l@{\hspace{0.6em}}l@{}}
AdamW & Adaptive moment estimation with weight decay \\
AI & Artificial intelligence \\
AND & Logical conjunction, true only if all inputs are true \\
argmax & Index of the largest value \\
B, M, K & Billion, million and thousand \\
bfloat16 & Brain floating-point format, 16 bits \\
CUDA & Compute Unified Device Architecture \\
E0--E3 & Short architecture-selection training runs (Table~\ref{tab:ablate}) \\
FLOPs & Floating-point operations \\
GB & Gigabyte \\
G1, G1b & Inner models of the 0.8B and 4B builds \\
GF(2) & Galois field of two elements (binary arithmetic) \\
KL & Kullback--Leibler (divergence) \\
kNN-LM & $k$-nearest-neighbour language model \\
L4 & NVIDIA L4 data-center graphics processor \\
LM$^2$ & Language model in a language model \\
MIT & Massachusetts Institute of Technology (license) \\
MLP & Multilayer perceptron \\
RASP & Restricted Access Sequence Processing language \\
relu & Rectified linear unit, $\mathrm{relu}(a)=\max(0,a)$ \\
RISC-V & Reduced instruction set computer, version five \\
top-1 & Accuracy of the highest-scoring character \\
XOR & Exclusive OR \\
\end{tabular}}

\section{Introduction}
A conversational language model can delegate a specific task to a smaller model trained for that task. Such delegation offers a potential route to more efficient inference. When a specialist is served remotely, delegation depends on communication with that service and its availability. Incorporating the specialist into the host's checkpoint offers an alternative: the component travels with the host and executes locally, without a separate specialist inference service. This could be useful for applications that must operate without remote-service access or seek to reduce the number of services they maintain. A separate specialist can also run locally, so embedding is a deployment option whose execution and storage costs must be assessed.

We study this deployment option using story generation. We call the larger conversational model the \emph{host} and the smaller story generator the \emph{inner model}. The inner model predicts and generates one character at a time. LM$^2$ is the system that embeds this generator into the host. Our question is whether designing the generator for direct compilation can reduce internal execution cost, and how this design affects its ability to predict text. We use story generation to measure that trade-off, without assuming that the generator writes better stories than the host.

Transformer weights can already encode programs: RASP describes transformer computation as a programming language \cite{weiss2021}, Tracr compiles such programs into weights \cite{lindner2023}, and looped transformers can emulate general-purpose computation \cite{giannou2023}. More recently, Qwenputer \cite{agrawal2026b} compiled a 32-bit RISC-V processor into additional layers of Qwen3.5-4B \cite{qwen2026b}, and QwenPU \cite{agrawal2026a} added a vector unit to it. This processor can execute the 260K-parameter stories260K language model \cite{karpathy2023} as a C program.

In QwenPU, the processor executes the inner model as a sequence of instructions. A \emph{host token} advances this internal execution, and it need not correspond to a character in the final story. Producing a single output can therefore require many host tokens. QwenPU reports approximately 53{,}000 host tokens per stories260K token \cite{agrawal2026a}, or approximately 25{,}000 per character using the measured 2.11 characters per stories260K token. To make the embedded specialist less costly to execute, we target the trained network itself for compilation. If the host layers evaluate all operations of one inner-model step directly, a single host forward pass can produce the next character.

Exact arithmetic is the central obstacle to direct compilation. The host uses bfloat16 \cite{kalamkar2019}, which represents every integer in $[-256,256]$ exactly but cannot represent every larger integer. Unrestricted neural-network sums can exceed this range and require additional arithmetic circuitry. LM$^2$ addresses this constraint by training the small model to use restricted integer calculations and simple on/off neurons. This allows each hidden layer to run exactly inside one feed-forward calculation block (MLP) of the larger host model. Hashed induction memory supplies context beyond the short character window.

Our pipeline trains the inner model, exports a fixed integer implementation, compiles its calculations into added layers, merges those layers with the host, and routes story requests to them. The exported implementation serves as the reference for checking the compiled model's outputs. With Qwen3.5-0.8B as host, the compiled component occupies 40 added layers. A circuit-only fast path, which runs only this component and skips Qwen's original layers while a story is being written, reaches 85--90 characters per second on one NVIDIA L4. We assess text prediction using bits per character, a loss measured on held-out stories for which lower values indicate better predictions. The LM$^2$ generator scores 1.145, compared with 0.894 for stories260K. Thus, the two systems trade fewer host tokens per character against the predictive performance of their different inner models.

Our contributions are:
\begin{itemize}
\item \textbf{Direct compilation of a trained language model.} We design and train a small language model whose calculations can be reproduced exactly inside a larger host model, combining both components in one checkpoint. The compiled model performs next-character prediction, memory updates, and character selection using one host token per generated character, and the host invokes it as an internal story-generation tool.
\item \textbf{Reduced storage requirements.} We reduce the space needed to hold intermediate calculations while the inner model runs. Together with preparing memory lookups only when needed, this reduces the complete checkpoint with the Qwen3.5-0.8B host from 22 to 18.1 GB.
\end{itemize}

\section{Related Work}
\textit{Compiled transformers:} RASP \cite{weiss2021} and Tracr \cite{lindner2023} map programs to transformer weights, and looped transformers execute instruction sets \cite{giannou2023}. Learning Transformer Programs works in the opposite direction, training a constrained transformer that can afterwards be converted into a readable program \cite{friedman2023}. QwenPU \cite{agrawal2026a} and Qwenputer \cite{agrawal2026b} compile a processor into Qwen3.5-4B, qwenGPU \cite{vixhal2026a} a processor that executes instructions across multiple parallel lanes into Qwen3.5-2B, and qwenFS \cite{vixhal2026b} a file system. LM$^2$ extends the QwenPU compiler to embed the calculations of a trained network directly. The inner network is trained to satisfy the host's numerical constraints so those calculations can be reproduced exactly.

\textit{Low-bit networks:} Binarized networks are trained through straight-through estimators \cite{bengio2013}, \cite{hubara2016}, and BitNet trains language models with binary and ternary weights \cite{wang2023}, \cite{ma2024}. Differentiable logic gate networks relax binary gates during training and discretize them afterwards \cite{petersen2022}, so that, like the threshold units here, the final model is exact by construction. For LM$^2$, we choose each layer's width and the number of nonzero inputs per neuron to keep its compiled calculations exact in the host's numerical format.

\textit{Memory:} Induction heads copy the token that followed an earlier occurrence of the current context \cite{olsson2022}. Outside the network, cache models reuse recently seen context \cite{grave2017}, kNN-LM retrieves next tokens from a datastore of earlier contexts \cite{khandelwal2020}, and infini-gram predicts from the longest matching $n$-gram of a large corpus \cite{liu2024}. LM$^2$ uses a fixed-size, hashed form of context retrieval: it predicts from the most recent continuation of the current context within one story, and it is stored in Gated DeltaNet state \cite{yang2025}, which Qwen3.5 interleaves with attention \cite{qwen2026a}.

\textit{Tools:} Language models can learn to call external tools \cite{schick2023}, whose results return through the host's text. LM$^2$ instead compiles the learned component into added host layers, with its output entering the context as ordinary tokens. This removes the need to serve that component separately, while retaining a host-side routing and decoding program. We evaluate the embedded implementation rather than comparative deployment costs against external tools.

\section{Proposed Method}

\begin{figure*}[t]
\centering
\includegraphics[width=\textwidth]{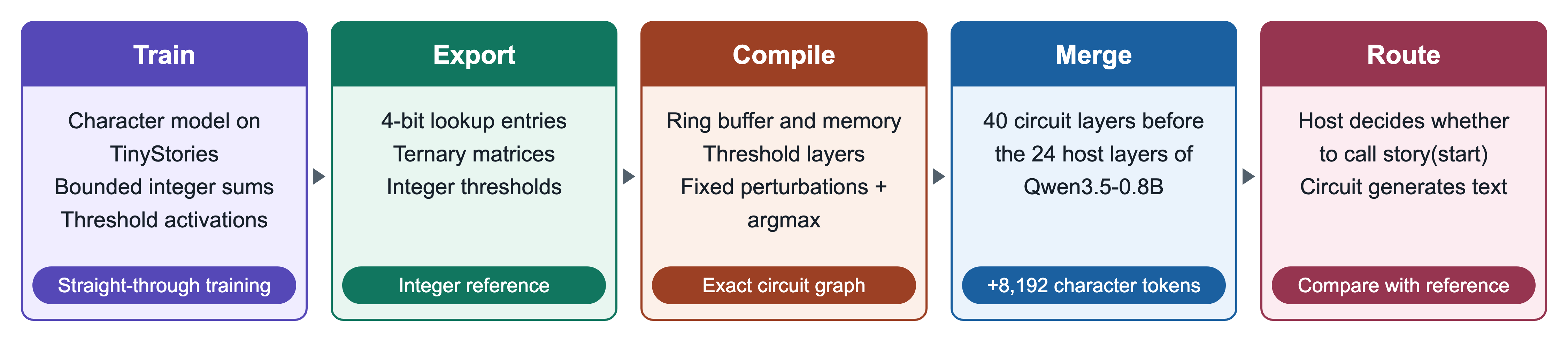}
\caption{The five stages of LM$^2$. Training enforces the arithmetic constraints, export defines the integer reference, compilation and merging embed its computation into the host, and routing invokes generation. Reference comparison is an evaluation check, separate from the router.}
\label{fig:pipeline}
\end{figure*}

To combine the host and specialist in one checkpoint, LM$^2$ transforms an arithmetic-constrained character model into a callable circuit inside a pretrained host. The compiled layers are added to the host, and the two components do not share the same learned parameters. Fig.~\ref{fig:pipeline} shows the five stages in execution order. Training defines the model and its memory, export fixes an integer reference, compilation maps the reference to circuit operations, merging embeds those operations into the host, and routing invokes character generation.

\subsection{Train: Arithmetic Constraints}
\label{sec:train}
The training constraint makes each hidden neuron a binary decision: it outputs 1 if its input sum $z$ reaches a threshold $\theta$, and 0 otherwise. The notation $[z \ge \theta]$ denotes this decision. Here $\ell$ and $u$ are the lower and upper bounds on $z$. For an integer sum $z \in [\ell, u]$,  the QwenPU compiler \cite{agrawal2026a} implements the decision with two MLP units:
\begin{equation}
[z \ge \theta] = \mathrm{relu}(z - \theta + 1) - \mathrm{relu}(z - \theta),
\label{eq:step}
\end{equation}
The rectified linear function is $\mathrm{relu}(a)=\max(0,a)$ for a scalar input $a$. The decision can also be implemented with the mirrored form for $[z < \theta]$. Every intermediate value must be an integer of magnitude at most 256 to stay exact in bfloat16, so either $u - \theta + 1$ or $\theta - \ell$ must be at most 256. A range span $u - \ell \le 511$ guarantees one of them. The inner model is therefore designed so that every hidden unit is $h = [z \ge \theta]$ for an integer pre-activation $z$ whose span is at most 511. Under this condition, one MLP stage evaluates a whole hidden layer without additional stages for propagating arithmetic carries.

\subsubsection{Character Model}
We call our final trained character-level story generator G1. G1 is the inner model that we compile into the conversational host, and the label identifies this model configuration. It predicts the next character over a 64-symbol alphabet. Table~\ref{tab:model} lists its architecture. We denote the number of neurons in hidden layer $r$ by $H_r$: G1 uses $H_1=4{,}096$, $H_2=2{,}048$, and $H_3=256$. Layer 1 reads the last $W = 16$ characters and the outputs of $M = 8$ memory heads, each encoded as a vector with one active entry (a one-hot vector), and sums one entry per input from a 4-bit table $T_1 \in [-7, 7]$:
\begin{equation}
z^{(1)}_j = \sum_{q=0}^{W+M-1} T_1[j, q, c_q], \qquad h^{(1)}_j = [z^{(1)}_j \ge \theta^{(1)}_j].
\label{eq:l1}
\end{equation}
In (\ref{eq:l1}), $q$ indexes the $W+M$ input slots, $c_q$ is the symbol in slot $q$, and $j$ indexes a neuron. The table entry $T_1[j,q,c_q]$ supplies that input\textquotesingle s contribution to neuron $j$, while $z^{(1)}_j$, $\theta^{(1)}_j$, and $h^{(1)}_j$ are its input sum, threshold, and binary output. The superscript $(1)$ denotes layer 1. The window length $W$ counts characters, whereas $H_1$ counts first-layer neurons. The span is at most $2 \cdot 7 \cdot 24 = 336$. Layers 2 and 3 multiply binary inputs by ternary weights, which take values $-1$, 0, or 1. Each row has at most 500 nonzero entries, so their spans are at most 500. The output scores, or logits, $L_c$ are ternary combinations of the 256 units of layer 3 plus an integer bias chosen so that $|L_c| \le 128$ for every possible input, and the next character is $\arg\max_c L_c$, ties going to the lowest index.

\begin{table}[tb]
\caption{Architecture of G1, the final character-level story generator compiled into the host. Width is the number of input slots or units at each stage.}
\label{tab:model}
\centering
\footnotesize
\begin{tabular}{@{}P{0.24\columnwidth}P{0.2\columnwidth}P{0.46\columnwidth}@{}}
\toprule
Stage & Width & Weights \\
\midrule
Inputs & $16 + 8$ & last 16 characters, 8 memory reads \\
Layer 1 & 4{,}096 & table, integers in $[-7, 7]$ \\
Layer 2 & 2{,}048 & ternary, $\le 500$ nonzeros a row \\
Layer 3 & 256 & ternary, $\le 500$ nonzeros a row \\
Output & 64 & ternary, integer bias, $|L_c| \le 128$ \\
\midrule
Total & & 5.7M nonzero integer weights \\
\bottomrule
\end{tabular}
\end{table}

\begin{figure}[tb]
\centering
\includegraphics[width=\columnwidth]{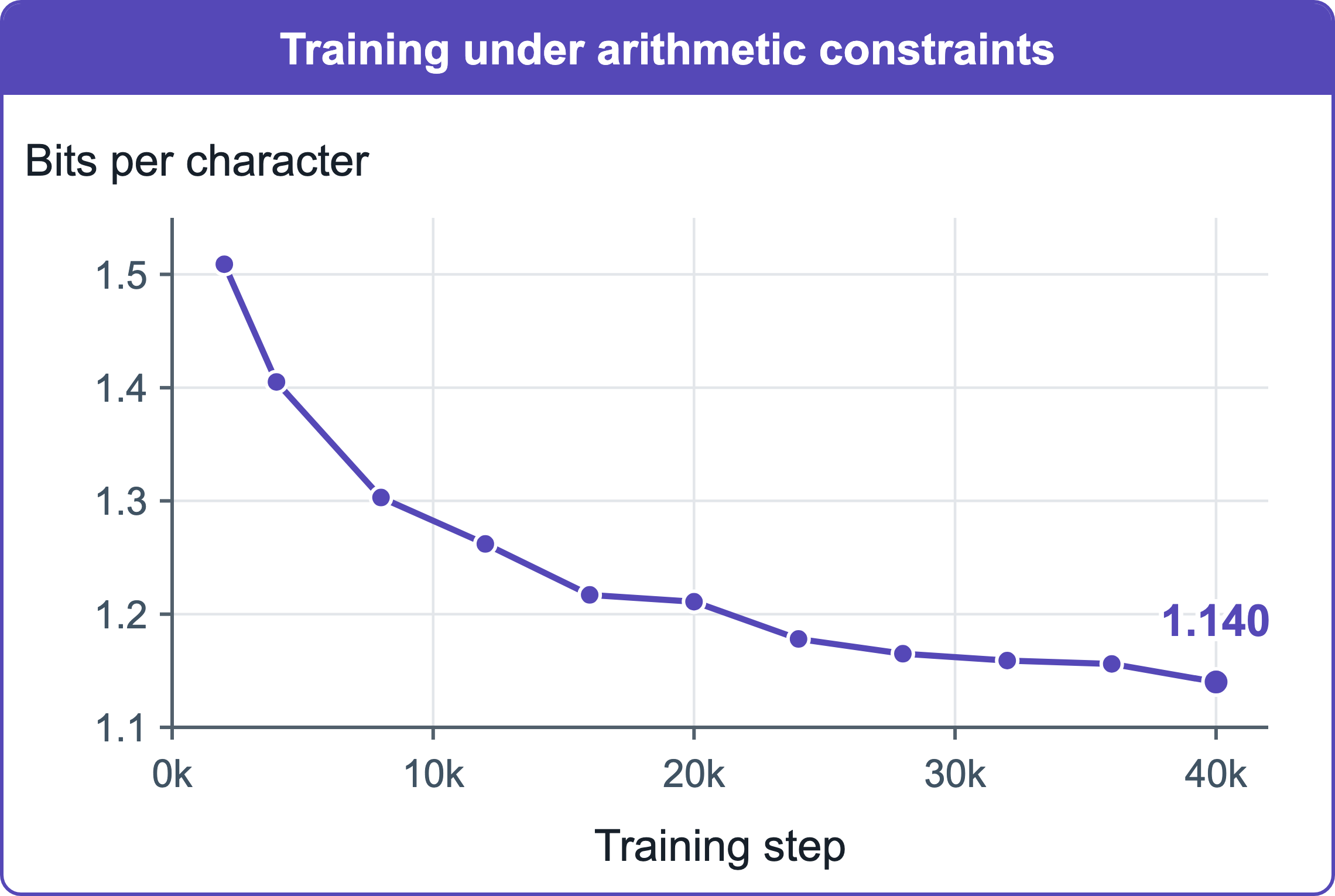}
\caption{Validation loss of the final inner model (G1) during training, from the integer-constrained float model.}
\label{fig:train}
\end{figure}

During training, weights are quantized in the forward pass and updated through straight-through estimators \cite{bengio2013}, with the straight-through gradient restricted to step-function inputs satisfying $|x| \le 1$, where $x$ is the input to the step function \cite{hubara2016}. Batch normalization \cite{ioffe2015} precedes each step function. Export converts these trained parameters into integer thresholds, as described in Section~\ref{sec:export}. The model was trained with AdamW \cite{loshchilov2019}, learning rate $2\times10^{-3}$ on a cosine schedule \cite{loshchilov2017}, batch 4{,}096 and 40{,}000 steps over 50 million characters of TinyStories \cite{eldan2023}, which took 73 minutes on one NVIDIA L4. Fig.~\ref{fig:train} shows its validation loss.

\subsubsection{Induction Memory}
The 16-character input window gives the model no direct access to earlier parts of a story. We add induction memory so that, when a short context reappears, the model can retrieve the character that previously followed it, even if that occurrence lies outside the window. Each memory head $i$ keeps an explicit induction table \cite{olsson2022} of the form ``which character followed this context the last time''. Here $i$ indexes a memory head and $k_i$ is its context length in characters. The head maps the last $k_i$ characters to one of 2{,}048 storage locations, called buckets. This mapping uses a hash linear over binary arithmetic (GF(2)), formed by applying exclusive OR (XOR) to fixed random table entries for each character position, with $k_i \in \{3, 4, 5, 6, 8, 2, 10, 12\}$.

\begin{figure}[tb]
\centering
\includegraphics[width=\columnwidth]{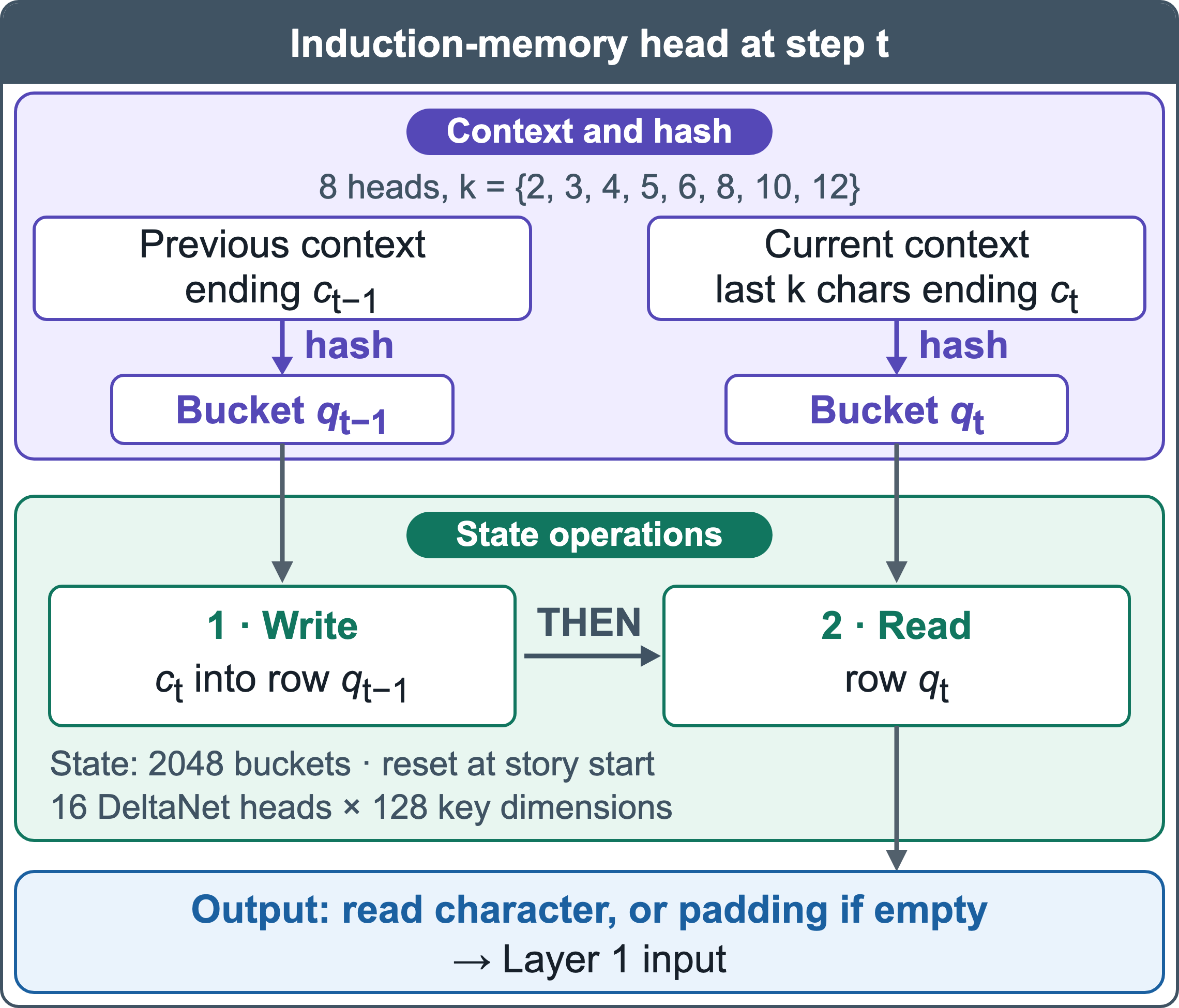}
\caption{One induction-memory head. Each step writes the current character into the bucket of the preceding context and reads the bucket of the current context. Eight heads use context lengths from 2 to 12.}
\label{fig:memory}
\end{figure}

Fig.~\ref{fig:memory} illustrates how one memory head records an observed continuation and retrieves a character for the next prediction. In the first operation (Write), the current character is stored in the bucket selected by the preceding context. In the second operation (Read), the head looks up the bucket selected by the current context. The retrieved character, or a padding symbol if that bucket is empty, is supplied to layer 1 as an additional input in (\ref{eq:l1}). The write occurs before the read, as indicated by the arrow between the two operations in the figure.

Different contexts can map to the same bucket, so a stored character can be overwritten by a different context. The reference tables reset at the start of each story. Section~\ref{sec:compile} describes their mapping to Gated DeltaNet state.

\begin{figure*}[t]
\centering
\includegraphics[width=\textwidth]{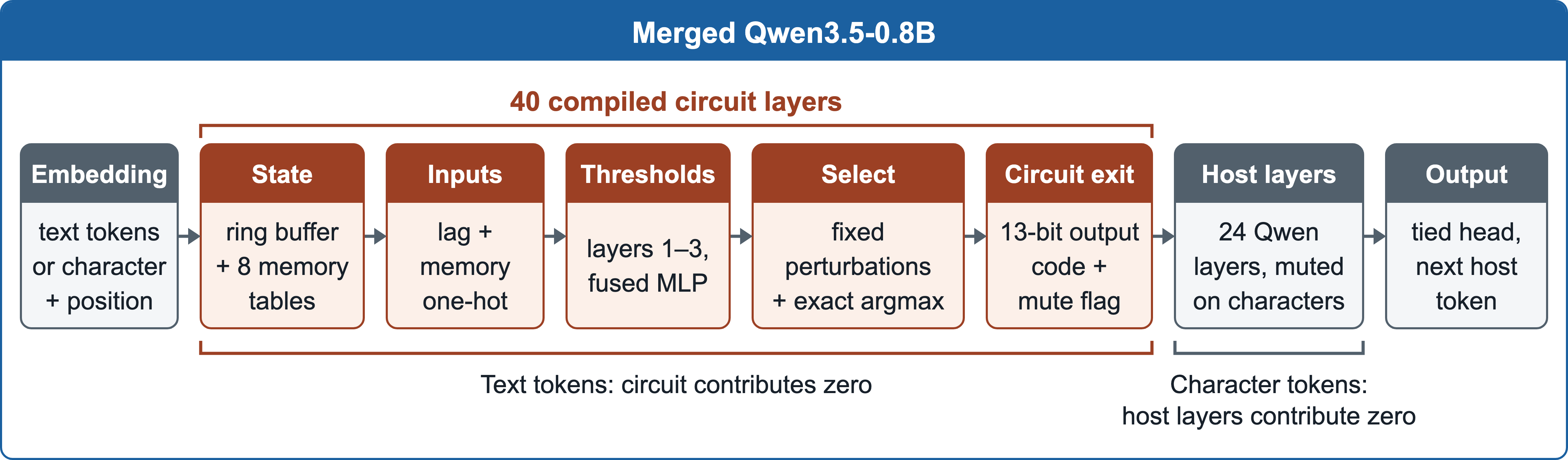}
\caption{Layout of the merged Qwen3.5-0.8B. On character tokens the circuit writes the next character's code and mutes Qwen's layers, while text tokens pass through it unchanged.}
\label{fig:layout}
\end{figure*}

\subsection{Export: Integer Reference}
\label{sec:export}
Export produces a fixed integer version of the trained inner model. This version is the reference against which we check compiled execution. The 4-bit lookup entries and ternary matrices become integer weights. Batch-normalization parameters and quantization scales are folded into one integer threshold per hidden neuron. The export also retains the memory configuration, output biases, and the fixed perturbation table used for generation. This reference defines exactness: a compiled sequence must match the exported model, rather than an unconstrained floating-point model.

\subsection{Compile: Stateful Generation Circuit}
\label{sec:compile}

Fig.~\ref{fig:layout} shows the merged model from left to right. The following paragraphs follow its named blocks, from Embedding through Circuit exit, and Section~\ref{sec:merge} explains Host layers and Output. The State through Circuit exit blocks together occupy the 40 compiled circuit layers.

\textit{Embedding:} Each inner-model token is a single host token that carries a character index $c$ and a position $p=t\bmod128$, where $t$ is the character-step index. The position cycles through 0--127, giving $64\times128=8{,}192$ new vocabulary entries. The embedding supplies one-hot vectors for the character and position. Ordinary text tokens use the host's text representation.

\textit{State:} The circuit retains recent characters in a ring buffer of four DeltaNet heads, whose convolution taps pack four characters into each row. The same block contains G1's eight induction-memory tables. Each table is stored in the recurrent state of Gated DeltaNet layers \cite{yang2025}. The gate remains open to retain stored values, and write strength 1 with one-hot bucket keys replaces the selected row rather than blending it with its old contents. The compiler evaluates each hash bit as a parity (XOR) over lag one-hots. Each table of 2{,}048 buckets uses 16 DeltaNet heads with 128 key dimensions each, and these implementation heads are distinct from G1's eight logical memory heads. At story start, the state decays to zero to reproduce the reference reset.

\textit{Inputs:} An MLP stage converts the stored character bits into one-hot vectors for the $W=16$ character lags, meaning the preceding 16 character positions. Together with the $M=8$ memory reads, these form the $W+M=24$ inputs in (\ref{eq:l1}). \textit{Thresholds:} The three hidden layers then apply the binary threshold decision in (\ref{eq:step}), each in a single MLP stage. The figure's ``fused MLP'' label refers to the handling of intermediate values described below.

\textit{Select:} The circuit chooses the character with the highest output score (argmax). For generation, it first adds the fixed perturbations described below. Selection uses $\binom{64}{2}=2{,}016$ pairwise comparisons $[L_a\ge L_b]$, where $a$ and $b$ index candidate characters and $L_a$ and $L_b$ are their scores. These comparisons are exact because $|L_a-L_b|\le256$. A 63-way AND combines a candidate's comparisons against the other 63 characters, with ties resolved by the lowest character index.

\textit{Circuit exit:} The winning character and next position are written as a 13-bit code: 6 bits identify one of 64 characters and 7 bits identify one of 128 positions. A mute flag makes the original host layers contribute zero on character tokens. The Output block converts this code into the next host token, as described in Section~\ref{sec:merge}.

\textit{Fixed perturbations in Select:} Greedy decoding often produces repeated text. To reduce repetition, the integer reference adds a fixed table of rounded Gumbel perturbations \cite{maddison2014}, indexed by position modulo 128 and character, before argmax. The perturbations are scaled using temperature 0.5 and clipped to $\pm16$ integer units, and training reserves this range by bounding the unperturbed logits by $128-16$. The compiler adds the table as a linear function of the position one-hot, without an additional circuit stage. Exactness here means exact reproduction of this fixed, quantized sampler. Because the perturbations are clipped and reused every 128 positions, this procedure does not provide independent Gumbel draws or exact categorical sampling.

\textit{Reducing intermediate storage:} The host passes a vector of intermediate values between its layers, called the residual stream. In the original compiler, each inner neuron's output occupies one entry in this vector until the next layer reads it. Following qwenFS's use of MLP intermediates \cite{vixhal2026b}, we extend the compiler so that one MLP unit can contribute directly to several sums in the next layer. This avoids storing the neuron's output in a separate residual entry before applying its outgoing weights. It is implemented by giving one MLP down-projection column multiple nonzero output rows. The sums are split into accumulators of at most 256 positive and 256 negative weights, so every partial sum stays exact. The 6{,}400 hidden values then never occupy the residual stream. Together with building each memory head's query one-hots just before that head reads them, this reduced the hidden size of the merged 0.8B model from 6{,}912 to 5{,}632 and the checkpoint from 22 GB to 18.1 GB.

\subsection{Merge: Host Integration}
\label{sec:merge}
\textit{Host layers in Fig.~\ref{fig:layout}:} Merging places the compiled circuit before the conversational host. Following the QwenPU merge \cite{agrawal2026a}, the circuit's 40 layers precede the 24 of Qwen3.5-0.8B \cite{qwen2026a}, using the host architecture's existing layer types, without requiring a new layer implementation. Qwen's matrices are zero-padded to the new hidden size and its normalization weights rescaled, so its computation is unchanged up to bfloat16 rounding. Every circuit node outputs exactly zero on text tokens, and a dedicated feature pushes character-token logits to about $-2^{20}$ at text positions. On character tokens the circuit's last layer keeps only the output code and sets a flag that mutes Qwen's layers, so they add nothing on character tokens and their DeltaNet recurrent state is preserved, although Qwen's attention layers still store keys for those positions and its short convolutions see them.

\textit{Output in Fig.~\ref{fig:layout}:} The tied output head shares weights with the embedding. It scores the character-position token matching the circuit\textquotesingle s 13-bit code highest. Greedy host decoding therefore returns the next character token selected by the integer reference.

\textit{Circuit-only fast path:} The merged checkpoint can be run in two ways. Normally every token passes through all 64 layers. While a story is being written, however, every new token is a character token, and on those tokens Qwen's original layers add exactly zero, so skipping them leaves the output unchanged. The fast path therefore runs each story character through only the embedding, the 40 compiled layers, the final normalization and the character rows of the output head. As in Qwenputer's chat \cite{agrawal2026b}, it records this sequence of operations once as a CUDA graph and replays it for every character. It uses the same weights as the merged checkpoint, ordinary chat turns still run through the full model, and each generated character is checked against the integer reference.

\subsection{Route: Internal Tool Invocation}
Routing decides whether the host should answer a message itself or delegate story generation to the inner model. Following qwenGPU's routing approach \cite{vixhal2026a}, the host is prompted to decide whether the message requests a story. The routing program compares the host's scores for ``yes'' and ``no''. If ``yes'' has the higher score, the host is directed to begin a story-generation tool call and writes the opening sentence. The inner model then continues the story one character at a time within the same conversation context. Otherwise, the host answers the message itself. The external program manages routing and decoding but does not generate story content.

\section{Experimental Setup}
The main evaluation uses Qwen3.5-0.8B as the conversational host and G1, the character-level model defined in Section~\ref{sec:train} and Table~\ref{tab:model}, as the embedded story generator. We examine text prediction, agreement with the integer reference, changes to the host, and routing. Table~\ref{tab:merge} also reports a Qwen3.5-4B implementation \cite{qwen2026b} with G1b, a character-model variant that continues G1's training from 40{,}000 to 80{,}000 steps and contains 5.5 million nonzero integer weights. This provides a comparison with QwenPU at the same host size, although the different inner networks prevent isolating the effect of host size alone. The processor-based comparison assesses internal execution cost, and no experiment compares deployment latency or resource use against a separately served specialist.

\textit{Text prediction:} We compare the LM$^2$ inner model with stories260K \cite{karpathy2023}, the model executed by QwenPU. Both are evaluated on the same 2{,}000 TinyStories \cite{eldan2023} validation stories (1.40 million characters). Bits per character measures the negative log probability assigned to the held-out text, normalized by character count, so lower values indicate better predictions. For G1 these values come from the constrained floating-point model whose integer export is compiled, and for stories260K from its original float32 checkpoint, whereas QwenPU executes stories260K with int4 weights. This measures text prediction rather than human judgments of story coherence. On the normalized validation text, stories260K tokens average 2.11 characters, while our model generates individual characters. Architecture-selection losses use a separate training-script validation sample and are reported separately.

\textit{Execution agreement:} We compare compiled calculations and outputs with the exported integer reference. Checks cover the circuit graph with every node rounded to bfloat16, 300-character runs of the merged checkpoint both alone and after conversational text, and every character of the tool demonstration.

\textit{Host behavior:} We compare the host before and after adding the compiled layers. Over 16 prompts, we measure Kullback--Leibler (KL) divergence between their next-token distributions, where a smaller value indicates a smaller distribution change. We also count identical 32-token continuations under greedy decoding, which always selects the highest-scoring token. Routing is checked on 60 messages, 30 story requests and 30 other requests, by comparing the host's scores for ``yes'' and ``no''. The four-message demonstration additionally runs the full tool invocation and replies.

\section{Experimental Results}

\subsection{Training Choices}
The labels E0--E3 identify short architecture-selection runs. In Table~\ref{tab:ablate}, $H_1$ denotes the number of neurons in the first hidden layer, and Memory denotes the number of induction-memory heads. G1 denotes the final model defined in Table~\ref{tab:model}. Table~\ref{tab:ablate} separates the short architecture-selection runs (60{,}000 steps, 50 million characters) from the final G1 configuration trained for 40{,}000 steps. The values use the training script's validation sample, rather than the shared 2{,}000-story comparison set. Adding eight memory heads reduced validation loss from 1.379 to 1.274 bits per character at the same layer width. Widening the first layer reduced it further to 1.236. Sixteen heads did not improve on eight, while reducing each table to 512 buckets increased loss by 0.033 bits per character. Without memory, a 32-character window did not help, since layer 1 only adds per-lag contributions.

\begin{table}[tb]
\caption{Architecture comparisons on the training-script validation sample. $H_1$ is the first hidden layer\textquotesingle s neuron count, and Memory is the number of induction-memory heads. Top-1 is the fraction of correctly predicted next characters. E0--E3 are short-run comparisons, while G1 is the final configuration, with a separate training schedule. These values are distinct from the shared 2{,}000-story evaluation.}
\label{tab:ablate}
\centering
\footnotesize
\setlength{\tabcolsep}{3pt}
\begin{tabular}{@{}lccc@{}}
\toprule
Run & Memory & Bits/char & Top-1 \\
\midrule
E0, $H_1 = 1{,}024$ & 0 & 1.379 & 0.719 \\
E1, $H_1 = 1{,}024$ & 4 & 1.280 & 0.741 \\
E2, $H_1 = 1{,}024$ & 8 & 1.274 & 0.738 \\
E3, $H_1 = 2{,}048$ & 8 & 1.236 & 0.745 \\
\midrule
G1 (final, Table~\ref{tab:model}) & 8 & 1.140 & 0.765 \\
G1, collision-free memory & 8 & 1.128 & -- \\
\bottomrule
\end{tabular}
\end{table}

Variants whose memory avoids hash collisions reached 1.128 bits per character with eight heads and 1.125 with sixteen, compared with 1.140 for G1 with hashed memory. These variants were trained but not compiled, so their losses are not results from an embedded model.

\subsection{Sampling and Repetition}
\begin{figure}[tb]
\centering
\includegraphics[width=\columnwidth]{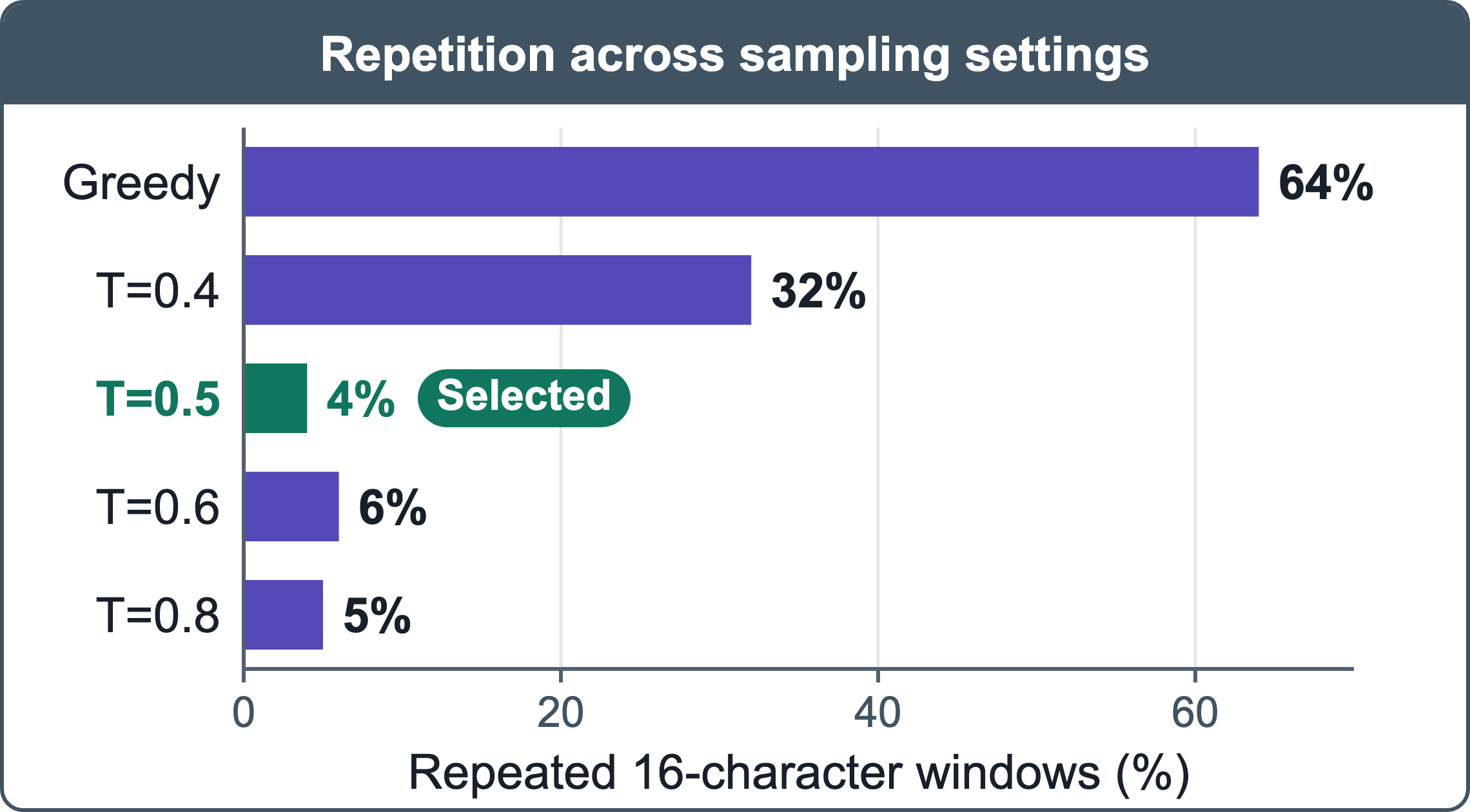}
\caption{Repeated 16-character windows in sampled text, by sampling temperature of the in-circuit Gumbel table.}
\label{fig:rep}
\end{figure}

Fig.~\ref{fig:rep} shows the share of 16-character windows that repeat earlier in the same sample (three prompts, 400 characters each). With no perturbations (greedy decoding), 64\% of windows repeat. Scaling the fixed perturbation table with temperature 0.5 reduces this share to 4\%. Higher temperatures give noisier text in these samples, so the demonstration uses 0.5.

\subsection{Exactness and Cost to the Host}
Table~\ref{tab:merge} reports the merged Qwen3.5-0.8B implementation. It matches G1's integer reference in the circuit checks and the 300-character generation checks, alone and after text. The mean KL divergence from the base host is $6.65\times10^{-4}$ over the 16 evaluated prompts, and 15 of 16 greedy continuations are identical to the base model's.

\begin{table}[tb]
\caption{Main 0.8B implementation and additional 4B implementation. Each is checked against its own integer reference. Host-output and routing checks use 16 prompts and 60 messages, and bits per character uses the shared 2,000-story set.}
\label{tab:merge}
\centering
\footnotesize
\setlength{\tabcolsep}{3pt}
\begin{tabular}{@{}P{0.46\columnwidth}P{0.23\columnwidth}P{0.23\columnwidth}@{}}
\toprule
Property or check & 0.8B + G1 (main) & 4B + G1b \\
\midrule
Compiled + host layers & 40 + 24 & 16 + 32 \\
Hidden size (base) & 5{,}632 (1{,}024) & 11{,}776 (2{,}560) \\
Checkpoint & 18.1 GB & $\sim$56 GB \\
Host tokens per character & 1 & 1 \\
Circuit graph matches reference & yes & yes \\
300 characters match (alone, after text) & yes, yes & yes, yes \\
Mean KL(base $\Vert$ merged) & $6.65\times10^{-4}$ & $6.90\times10^{-4}$ \\
Greedy continuations identical & 15/16 & 11/16 \\
Router accuracy (60 messages) & 93\% & 100\% \\
Bits per character & 1.145 & 1.132 \\
\bottomrule
\end{tabular}
\end{table}

The compiled operations occupy 40 layers preceding the 24 layers of the Qwen3.5-0.8B host.

\subsection{Tool Routing}

We tested whether the host's routing decision would delegate story requests to the inner generator and keep other requests. The test contained 30 story requests and 30 other requests. Qwen3.5-0.8B's yes and no scores selected the tool for all 30 story requests and for 4 of the other requests, giving 56 correct routing decisions out of 60 (93\%).

The four errors were requests that the host incorrectly sent to the story generator. These asked for a haiku, a limerick, information about the history of the Eiffel Tower, and an answer about a story the user had previously read.

\subsection{Comparison with the Program-Based Approach}
Table~\ref{tab:compare} compares two ways of executing a language model internally: QwenPU \cite{agrawal2026a} runs stories260K \cite{karpathy2023} as a program, while LM$^2$ directly compiles its own trained character model. The comparison reports host-token count, checkpoint size, and text-prediction loss. The systems execute different inner networks and use different host sizes. QwenPU requires about 25{,}000 host tokens per character, while the Qwen3.5-0.8B implementation of LM$^2$ requires one. This token-count ratio is not a wall-clock speedup: the builds differ in hidden width, circuit depth, and execution mode. The 18.1 GB checkpoint is smaller than QwenPU's 26.7 GB checkpoint, but it also uses a smaller host. At the same 4B host size, LM$^2$ still requires one host token per character, but its approximately 56 GB checkpoint (Table~\ref{tab:merge}) exceeds QwenPU\textquotesingle s 26.7 GB.

\begin{table}[tb]
\caption{QwenPU and the main LM$^2$ implementation. Host and inner models differ, and lower bits per character indicates better text prediction.}
\label{tab:compare}
\centering
\footnotesize
\setlength{\tabcolsep}{3pt}
\begin{tabular}{@{}P{0.36\columnwidth}cc@{}}
\toprule
Property or metric & QwenPU & LM$^2$ \\
\midrule
Host & Qwen3.5-4B & Qwen3.5-0.8B \\
Inner model & stories260K & G1 \\
Inner weights & 260K int4 & 5.7M int. \\
Host tokens per character & $\sim$25{,}000 & 1 \\
Checkpoint & 26.7 GB & 18.1 GB \\
Bits per character & \textbf{0.894} & 1.145 \\
\bottomrule
\end{tabular}
\end{table}

Generation rates are reported for this work only, because the published QwenPU rate comes from a dedicated executor that runs its circuit layers alone \cite{agrawal2026a}. On one L4, the 0.8B build's fast path generates 85--90 characters per second. When every character passes through the full merged model, the rate was about 7 characters per second for an earlier 22 GB 0.8B build on an L4.

The lower host-token count accompanies weaker text prediction in our model. On the shared 2{,}000-story set, stories260K scores 0.894 bits per character and the LM$^2$ G1 model scores 1.145 (Table~\ref{tab:compare}). Lower values are better. For G1, both the constrained floating-point implementation and the integer implementation achieve 0.760 top-1 accuracy, meaning their highest-scoring next character is correct at 76.0\% of evaluated positions. This accuracy comparison concerns two implementations of G1, not G1 versus stories260K.

\subsection{Compute per Host Token}
Each character token executes one full step of the inner model, about 1.26 million integer weight operations: 98 thousand in layer 1 including the memory inputs, 1.02 million in layer 2, 128 thousand in layer 3 and 6 thousand in the output layer. For QwenPU, a host token represents about six weight operations of stories260K, by our estimate from its 53{,}000 host tokens per stories260K token. This estimate describes inner-model operations represented per host token. It does not measure host FLOPs, energy efficiency, or wall-clock throughput, because lookup additions and floating-point multiply-adds have different execution costs.

\subsection{Tool-Use Demonstration}
Table~\ref{tab:demo} reproduces the recorded demonstration on Qwen3.5-0.8B. Qwen answered the two factual questions itself and called the tool for both story requests, writing each story's first sentence inside the call. The inner model then continued for 300 characters, identical to the integer model.

\begin{table}[tb]
\caption{Recorded Tool-Use Demonstration on Qwen3.5-0.8B}
\label{tab:demo}
\centering
\footnotesize
\setlength{\tabcolsep}{3pt}
\begin{tabular}{@{}P{0.26\columnwidth}P{0.12\columnwidth}P{0.53\columnwidth}@{}}
\toprule
User message & Answer & Reply (beginning) \\
\midrule
Who are you? Answer in one sentence. & Qwen & I am an AI assistant designed to help you with various tasks. \\
Tell me a story about a girl named Lily. & tool & Once upon a time, there was a little girl named Lily. She loved to play outside. One day, a big dog and a sick. \ldots \\
Can you tell me a bedtime story about a dog? & tool & Once upon a time, there was a little dog named Barnaby. He wanted to find a new toy in his mouth. \ldots \\
What is the capital of France? & Qwen & Paris. \\
\bottomrule
\end{tabular}
\end{table}

\section{Discussion}
\textit{Deployment implications:} LM$^2$ packages the conversational host and a learned generation component in a single checkpoint. Once loaded locally, the component requires no separate specialist inference service. This arrangement motivates studying embedded capabilities for applications without remote-service access. It is not unique to offline operation: a separate small model can also run locally, can be updated independently, and may have lower resource requirements. Our results establish an execution mechanism and its costs, rather than a deployment advantage over that alternative.

\textit{Inference implications:} In the circuit-only fast path, story generation bypasses the host's original layers. This shows how internal delegation can avoid executing those layers during the specialist's turn. The reported 85--90 characters per second measures that path, and there is no matched comparison against host-only story generation or a separately executed small model. The host-token reduction relative to QwenPU also compares different inner models, so it represents a design trade-off rather than an acceleration of the same network.

\textit{Predictive capability:} Restricting the inner model during training lets the compiler implement each hidden layer in one host MLP stage. The resulting models predict text less well than stories260K. On the training-script validation sample, wider layers, collision-free memory, and continued training to 80{,}000 steps (1.123 bits per character) each provided only modest improvements after memory was added. On the separate shared validation set, G1 scores 1.145 bits per character versus 0.894 for stories260K, a gap of 0.251 bits per character. Integer thresholds, a fixed 16-character window and per-lag lookups are plausible constraints on expressiveness, although the experiments do not isolate their individual effects. A compiled processor, by contrast, can run any network that fits its memory, at a much higher cost per character.

\section{Conclusion and Limitations}
\label{sec:limits}
LM$^2$ investigates self-contained deployment of a conversational host and a learned specialist by addressing the cost of internal neural-network execution. Training the inner model for direct compilation enables one next-character step per host token, compared with approximately 25{,}000 for QwenPU running stories260K. The 18.1 GB checkpoint combining G1 with Qwen3.5-0.8B invokes the character model as an internal story-generation tool and matches its integer reference in the reported checks. This reduction in host-token cost comes with lower predictive quality: 1.145 bits per character for the G1 generator versus 0.894 for stories260K on the shared validation stories, where lower values indicate better text prediction. The study demonstrates a feasible embedded implementation and quantifies its trade-offs, but it does not establish faster or cheaper inference than a separately served specialist.

\textit{Model and task scope:} The main evaluation concerns one threshold-based inner model, G1, embedded in Qwen3.5-0.8B for children's stories. The additional 4B implementation uses a second host configuration and the G1b variant, which continues G1's training. Generated text loses entity consistency within a few sentences. The results do not establish generalization to larger inner models, other tasks, or unrestricted neural architectures.

\textit{Evaluation coverage:} Host-distribution checks use 16 prompts, routing uses 60 messages, and merged exactness checks cover 300-character runs. Agreement on these tests complements the arithmetic construction but is not exhaustive verification of every possible sequence. The 18.1 GB build's full-model throughput was not measured, and the fast-path rate cannot establish an end-to-end speedup over the processor-based reference.

\textit{Storage and sampling:} The added hidden width makes an 18.1 GB checkpoint large relative to the 0.8B host. A smaller checkpoint than QwenPU is demonstrated only by changing to this smaller host, since the 4B build in Table~\ref{tab:merge} is larger. The fixed perturbation table is deterministic and periodic, and does not implement independent categorical draws. Improving predictive quality while retaining exact execution and reducing checkpoint overhead are important steps toward practical embedded specialists. Comparisons with local and remote specialist services would be needed to establish when this deployment arrangement is beneficial.

\section*{Acknowledgment}
This work builds on the QwenPU and Qwenputer compiler by Divyansh Agrawal, released under the MIT license. The authors acknowledge the use of AI-based tools to assist in drafting and editing the text. All content was reviewed and verified by the authors, who take full responsibility for this work.

\section*{Code Availability}
The training, compilation, merge and evaluation code, together with the trained integer models and run logs, is available from the authors on request.

\IEEEtriggeratref{4}

\end{document}